\input amstex
\input xy
\xyoption{all}
\documentstyle{amsppt}
\document
\magnification=1200
\NoBlackBoxes
\nologo
\hoffset1.5cm
\voffset2cm
\pageheight {16cm}


\bigskip

{\bf COGNITIVE NETWORKS: }

\medskip

{\bf BRAINS, INTERNET, AND CIVILIZATIONS}

\bigskip

{\bf Dmitrii  Yu.~Manin,}

\medskip

{\bf Yuri I.~Manin,}  Max--Planck--Institut f\"ur Mathematik, Bonn, Germany.

\bigskip

{\bf Abstract.} In this short essay, we discuss some basic features of cognitive
activity at several different space--time scales: from neural networks in the brain to
civilisations. One motivation for such comparative study is its heuristic value.
Attempts to better understand the functioning  of ``wetware'' 
involved in cognitive activities of central nervous system by comparing it with a
computing device have a long tradition. We suggest that comparison with Internet
might be more adequate.
We briefly touch upon such  subjects as encoding, compression, and Saussurean
trichotomy {\it langue/langage/parole} in various environments.

\bigskip


\medskip{


\bigskip

{\bf Introduction}

\medskip

In several recent research papers and surveys by neuroscientists (cf. [1], [29], and references therein), it was suggested that
cognitive functions of the brain are performed using not only, and perhaps
even not mainly, complex networks of interacting neurons ({\it connectionist view}), -- but also
on the level of individual, highly specialized neurons and their {\it intracellular} mechanisms.
This argumentation went hand in hand with the critique of popular analogies between brains and computers,
where neurons were supposed to work as, say, electronic logic gates.

\smallskip

In order to retain the heuristic power of computer science in cognitive neurobiology and simultaneously to
keep the door open to such  paradigm extension, we consider in this paper
possible analogies between the brain and Internet, in which certain neurons and some
specific neural networks are being
compared with entire computers, in particular, with servers, that in fact do have
a very rich internal hardware and software reflected in their functions in the net.

\smallskip

As so many basic ideas and technologies of the information age,
the future role of the World Wide Web was presciently understood
by Alan Turing. Although the Internet of course did not
yet exist then, according to a convincing interpretation by 
B.~Jack Copeland ([4], p.~30), Turing's definition and study of oracle machines in his
 PhD thesis (1938) introduced the notion that computability may involve getting
 ``oracular'' data from outside computers.

\smallskip

In this essay, we do not discuss any philosophical problems
related to such comparisons (for possibly related discussions, see e.~g.~[14], [3], and references therein). 

\smallskip

We simply try to suggest
plausible and verifiable conjectures about 
functions, interconnections, and dynamics  of various neural
structures in the brain using the brain/Internet metaphor. Comparison
with computer was already exploited in the enthusiastic book by Jeff Hawkins  [11] (cf. further
developments in  [12], [10], [13].)
\smallskip

Another subject matter concerning us here is the cognitive
activity of civilizations. Looking for cognitive network patterns at this level
is not a standard preoccupation of the historians of culture and sciences,
but one of us first engaged in this line of thinking when
researching available data on the development of writing, cf. [24].

\smallskip
Our departure point is
a simple remark: although WWW is very complex,
the knowledge about its structure and functions is available on all levels since
it is constructed and developed by means of engineering, cf. the first section below.
Contrariwise, brains are products of evolution,
we can observe their structure and functioning at various 
spatial and temporal scales, but we can only venture some guesses about 
their modes of data processing. AI might be and in fact was a great inspiration
for such guesses.

\smallskip

Important role  in this circle of ideas is played by the notion
of {\it information transmission}. Generally, we imagine a source
of data, which can be encoded,  transferred  as a message through a certain channel,
received at the other, and then decoded to reconstruct the initial data.

\smallskip

Using a Saussurean terminology, we can say that  protocols at the transmitter/receiver 
ends constitute a language {\it (la langue),} whereas each case of transmission
is an act of speaking {\it (la parole)}.
 
\smallskip

There are many mathematical models of information transmission 
materialized in the IT domain. What we want to stress in this note is the fact that
actual transmission must often be relayed: data at the receiving end become,
after
re--encoding, data at the sender along another channel, and so on.
This involves the rules and protocols of {\it translation} in
linguistics. Usually,  basic parts of such protocols are accessible as
bilingual dictionaries, but even in human societies
there are exotic exceptions such as {\it drum languages} of
various ethnic groups of Africa, New Guinea, etc.

\smallskip

Thus, there must exist many {\it neural languages}, each used in respective
neural networks, and connected by numerous translating neurons/networks.

\smallskip

Finally, the speed with which the brain can solve cognitive problems
related to speech generation and recognition (or such a marginal
activity as playing chess) unambiguously testifies to
the abundance of highly parallel processing in neural networks.  Neural organization of
such parallel processing must be a very essential logistical task.
This was long ago recognized and described by neuroscientists dealing
with mechanisms of visual perception.

\smallskip

Here we must stress
that mathematical theory of {\it time complexity of parallel processing}
practically does not exist, and in any case
did not reach maturity comparable with those
of {\it Kolmogorov complexity} and  {\it polynomial time computations.}

\smallskip
Therefore a better understanding of high parallelism in the brain
might serve as a useful heuristic tool for theoretical computer science as well:
cf. [17] and many other studies of visual cortex.

\smallskip

{\it Acknowledgements.}  Yu.~I.~Arshavski in his ample correspondence with Yu.~Manin
discussed and clarified for us various problems of modern neuroscience and the relevant problems
of AI. Earlier articles by and email exchanges with Nora Esther Youngs, Carina Curto, and Vladimir Itskov
were also very stimulating. We are cordially grateful to them.

\medskip

\centerline{***}

\smallskip

This essay is  organized as follows.
After a brief  survey of the global structure of WWW, we discuss
the following subjects:

\smallskip

$\bullet$ Architecture of WWW and the role of search engines.

\smallskip

$\bullet$ Chips, computers, and servers vs neurons and neural networks.

\smallskip

$\bullet$ Kolmogorov--style compression vs.  Charles Darwin--style compression.

\smallskip

$\bullet$ Miscellany.

\bigskip

{\bf Computer networks: Architecture}

\medskip

When describing the information pathways in computer networks, specifically, the Internet, one has to keep in mind that communication between network nodes (individual devices, be that special-purpose servers, personal computers or network-enabled electronic devices) can be considered on several different levels (cf. the OSI model, 
[31] and [31] [OSI1] where the respective levels are called {\it layers}). From the lowest level concerned with transferring of raw bits between two neighboring devices to the highest level that operates in terms of such operations as remote file access or search engine queries, each layer has its own semantics and serves as a medium for the next higher layer. On the lowest levels of the hierarchy, each device can communicate only with its immediate neighbors. On the intermediate levels, the complexities of networking are hidden, and nodes can directly address their requests to specific other nodes (identified by IP addresses). Finally, on the highest levels, the notion of a network node is also hidden, and they operate in terms of services, such as a named file share or a particular search engine. 

\smallskip

What makes this transparency possible is the existence of "routing protocols" encapsulated in a special class of Internet servers called {\it routers}. Without getting into details, routers keep exchanging information they learn about network nodes existing in their neighborhood. Since the network configuration keeps changing (as nodes come up on and drop off the network, as new subnets are added and old ones reconfigured), the routing information is never complete, always to some extent outdated, and often contradictory. The robustness of communication between nodes is only achieved by the routers' ability to retry delivery of lost messages using different routes. What is most interesting for us here is that even considered on this level, Internet possesses a (varying
with time)  map of itself, in particular of its own topology. This map is approximate, somewhat fuzzy, partially delayed, and decentralized. Perhaps it can be likened to the living organisms' proprioception. 

\smallskip

From the point of view of information processing, we should look at the application level of the OSI model. As in the brain, information enters the network through peripheral nodes, i.e., mostly consumer devices where people type in texts and upload images or videos. Some of it stays local, of course, but some of it travels on the network to be stored, transferred to other peripheral nodes (e.g. email) or, most interestingly, processed, digested, summarized, and transformed. We can discern several types of memory-like subsystems in the network. 

\smallskip

{\it 1. Storage systems.} These are places (server farms) built to provide the archival and backup functions to the users, such as DropBox or Google Drive. They are probably the least interesting type of network "memory," returning exactly what was put in there on a specific request to retrieve it.

\smallskip
{\it 2. Internet archives, such as the Wayback Machine.} It crawls the web and stores the current copies of the sites it visits, without overwriting the older versions. Thus, it allows one to reconstruct the history of  web's
dynamics, though in an unavoidably patchy form. 

\smallskip

{\it 3. Internet search engines.} Search engines started as simple keyword retrieval databases of the important information gleaned from the web  but have evolved into powerful associative memory-type services. What's most interesting about search engines is that they increasingly perform deep analysis of both the content they index and the search patterns of the users, attempting to serve ever more complex and fuzzy user queries. There is an understanding that to effectively respond to difficult informational queries, a search engine has to possess at least a rudimentary type of world knowledge, such as Google's Knowledge Graph 
 and similar systems developed by other search companies.
 
 \smallskip
 
  Note that a significant portion of world knowledge (perhaps, the vast majority) in such systems is harvested from the web, rather than being manually entered. Search engines perform many different analyses of the content they index, like news aggregation (do these two news articles talk about the same event? if so, who are the event actors?), sentiment analysis (is this a positive or negative news story?) or image recognition (what objects are in the photo?).

\smallskip
Search engines represent the kind of information storage that is inherently capable of self-reflection. As a rudimentary, but highly visible example consider several incidents where a search engine's algorithms would make a funny or offensive mistake in response to a query, which would become a news item, and then very soon the results it would return to the same search query would prominently feature news about its own mistake in what could be perceived as a form of self-deprecating humor.

\bigskip

{\bf Neural systems:}

\medskip

{\bf experiments, measurements, and self--perception}

\medskip

The only direct information channel to one's neural system
for each human being is self-perception, including 
memory, emotions, conscious sensory perceptions ("I see"
means "I know that I see").

\smallskip

Objective information about neural systems of other people,
but also of animals belonging to different species
is obtained in laboratories and clinics, but this is an outsider's
information.

\smallskip

Bridging together insider's and outsider's views has always been and remains a great
challenge. In particular, clinical and scientific interpretation of the 
data of psychology and psychiatry can be 
hopelessly caught in the trap
 of {\it suggestion}: cf. a very convincing study of the history
 of psychoanalysis in  [2].

\smallskip

Attempts of such bridging based upon computer metaphor were numerous.
Below we will briefly survey some of the conjectures summarized in Yu.~M.'s
paper of 1987 ``On early development of speech and consciousness (phylogeny)'',
see [18], pp.~169--189.

\smallskip

Basically, it was conjectured that the brain contains inside 
a map of itself, and that some neural information channels
in the central neural system:

\smallskip

{\it a) carry information about the mind itself, i.e.,  are reflexive;

\smallskip
b) are capable of modelling states of the mind different from the
current one, i.e. possess a modelling function;

\smallskip
c) can influence the state of the whole mind and through that the
behavior, i.e. possess controlling function.} ([18], p.~179).

\medskip

It was remarked also that that this reflection of the brain inside itself must be
unavoidably coarse grained. 

\smallskip

This is made much more precise in the  already invoked above 
 OSI (Open Systems Interconnection) models of the Internet,
where both the notion of the network node and protocols of their
communication are subdivided into ``horizontal'' layers (seven in [31] [OSI1]).
The lowest layer represents  the topology of physical medium 
 transmitting ``raw bit streams,'' whereas the highest layer 
 represents the most coarse-grained vision of the whole network.
 Each layer has its own communication language; each individual transaction
 (information transmission) on a particular layer can involve multiple
 transactions on the next lower layer and, in turn, serve as a part of a
 transaction on the next higher layer. Thus, information transmission of the
 highest layer data is mediated by multiple translations down to the lowest
 layer at the source, a corresponding translation up at the destination and
 potentially multiple partial up and down translations at the intermediate
 points. 

\smallskip

 We stress again that streams of bits on the wire directly
 represent only the lowest-level communication. In order to decode higher-level
 transactions, one would inevitably have to ascend the hierarchy of languages,
 aggregating multiple lower-layer conversations into a single higher-layer
 conversation: there is no way to directly jump from the lowest to the highest
 layer. The same is  true about the electrochemical messaging in the
 brain: individual trains of neuronal spikes do not directly represent thought or perception
 patterns.  This is of course well understood by experimental
 neuroscientists who use expression {\it "signature of \dots ''}
in articles summarizing their findings (cf.~ [16], [30]). 

\smallskip

As WWW, the mind can  contain several dynamical reflections of itself,
differently positioned with respect to the functions of mutual reflection
and control. The respective functional modes of the mind manifest
themselves in a wide variety of dissociative phenomena: multiple
personalities, automatisms, fugues, hypnotic phenomena, etc. 

\smallskip

Concrete implementations of fragments of multilayered structure in the brain are evident,
for example, in the studies of processing of sensory information of different
modalities. The way from a sensory input to the appropriate neural network 
in the respective projection area should be imagined as ``vertical'' information transfer
from lower layers up. On the other hand, integration of different modalities,
storage of the compressed form of this information etc. should
involve a considerable role of horizontal conversation.

\smallskip

In the human brain, anatomy of  neocortex
involves several (six) layers, and Jeff Hawkins made a series
of conjectures about storing and processing information
inside and between these layers (see [11], pp.~42,  237--245).
In Yu.~Arshavski's opinion (private communication), at least part of these conjectures
can be or have been experimentally verified, but the general association
of these anatomical layers with processing layers is hardly
justified.

\smallskip

We believe that understanding of such phenomena as  cognitive maps
of spatial environment [8], mirror neurons [9], or 
concept cells [27], [28], can benefit from a purposeful search
of WWW-like layers and decoding their languages (cf. [29], [1]).

\smallskip

Information about these layers might also enrich the current rigid
juxtaposition of ``purely connectionist'' paradigm and the 
``intracellular'' paradigm, according to which cognitive processes
are primarily served by chemistry and genetics of specialized cells
rather than by firing of individual neurons connected into networks.
It seems clear that memory  must involve chemistry and genetic
structures and cannot be based solely on network dynamics.

\bigskip

{\bf Information, compression, computation}

\medskip

{\bf Civilisational layer of cognition.} In [20], one of us argued
that cognitive processes in the human brain might and ought  to be theoretically considered
also at one level above the individial brain,
namely, on the {\it civilizational layer.} 

\smallskip

Nodes of this layer are individual brains but also,
starting with early modernity, it is enriched with libraries, laboratories,
research institutes, etc.

\smallskip

Comparison of this layer with (more formalized conceptually)  layers
involving primarily computers was based upon the following
suggestion. Let us focus on physics, science that  dominates
today our understanding of the universe along the vast spectrum of spatiotemporal scales.

\smallskip

It is a common knowledge that physics discovers ``laws of nature''
that are expressed by compact mathematical formulas.
These laws of nature can be then used for prediction/explanation of results
of observations (say, in astronomy) and of experiments, and also for engineering
projects.

\smallskip

It was suggested in [20] that each physical law might be considered as an {\it analogue
of a computer program.} Such a program computes the {\it output } after
accepting {\it results of observations as an input.} These outputs
are  {\it ``scientific predictions''}. The classical example consists in predicting
observable positions of planets using models by Ptolemy, Galileo, Newton, Einstein, etc.

\smallskip

This process might  also involve
other laws/programs, multiple relaying,
encoding and decoding that converge at an additional civilisation layer node, etc.

\smallskip

As a contemporary example, consider the recent  news that
the international team of scientists using
LIGO (Laser Interferometer Gravitational-Wave Observer)
was able to detect gravitational waves
and identify their source: two colliding black holes. 

\smallskip

Roughly speaking,
gravitational waves  resonate with light waves, because high-frequency
oscillations of space-time curvature (caused by gravity) cause the entire system of light-like 
geodesics (which in the first approximation determine the light propagation) to
oscillate at the same frequency.

\smallskip

The basic ``physical law'' involved in this event consists of
Einstein general relativity equations and its solutions
of a special type (black holes). 

\smallskip
At the node
of observations, a large sample of other ``physical laws''
is invoked that determine engineering decisions needed to construct
the big observational device called LIGO which detects very small 
frequency changes of laser beams using the interference techniques.

\smallskip

Finally, at all stages, actual computers are used, whose inputs and outputs 
represent ``vertical'' communication between an upper and a lower level
involved in this observational activity.

\medskip

{\bf Mathematical models in computer science: computability, complexity, polynomial
time.}
It is well known that the mathematical theory of computability was created
in the 1930s and 1940s  in several different versions: Turing machines  {\it (engineering metaphor)},
Church's lambda--calculus {\it (linguistic metaphor)}, Markov's algorithms
{\it (conveyor belt metaphor)}, Kolmogorov--Uspensky's algorithms {\it (information flow chart metaphor)},
partial recursive functions {\it (operadic metaphor)}  et al.

\smallskip

All these versions differ in many respects. First of all, their respective
domains of inputs and outputs  viewed as Bourbaki-style structures
are different: finite sequences of bits (zeros and ones) for a Turing
machine, finite words in an arbitrary fixed alphabet for a Markov's
algorithm, and words of a language which is the basic object of lambda--calculus.
Second, programs for particular computations are formalized
differently as well: a finite list of inner states of pairs {\it (head, head input)}
for a Turing machine; a finite word expressing the sequence of basic operations
on recursive functions together with their inputs etc.

\smallskip

Nevertheless, it was proved that all these constructions produce ``one and the same''
notion of computability, in a well-defined mathematical sense.
One of the most remarkable events in the nascent computer science
occurred when one of the founding fathers stated his famous  ``Church's thesis'':  {\it the computability notion 
is absolute and does not depend on the chosen model of computation}
(if the latter is broad enough). 

\smallskip

This thesis is {\it not} a mathematical theorem: it can be called an ``experimental fact
in the Platonic world of ideas.''

\smallskip

The next great discovery in this domain was that of   ``Kolmogorov complexity.''

\smallskip
If a model of computability and the suitable  programming language are chosen,
then one can prove the existence of the best compressing program $U$ with the following
properties. 

\smallskip

(a) Let $Q$ be an arbitrary object in the domain of this computability
model or else, a description of a partial recursive function, $U$ a semi-computable function.
Define the {\it complexity of $Q$ with respect to $U$} as the bit length
of the shortest object $P$ such that  $U(P)=Q$ (or, respectively, $Q$ is a program of computation
of the same function).   In other words, $P$ taken as input of $U$
produces $Q$ as its output. Such a $P$ always exists.

\smallskip

(b) There exists a class of optimal choices of $U$ such that a different choice 
of the universal programming language and/or  of 
another $U$ leads only 
to a possible change of complexity (as function of $Q$)
by a bounded additive constant.

\smallskip

Intuitively, this shortest object $Q$ is best imagined as a  {\it maximally compressed
form of $P$.}  Thus, we may say that Newton's classical laws of celestial
mechanics 
$$
F=G\frac{m_1m_2}{r^2}, \quad a = \frac{F}{m}
$$
are maximally Kolmogorov-compressed representations of programs
that can calculate and predict future positions of celestial bodies,
where observations of their current positions are taken as inputs.

\smallskip

Arguably, this Kolmogorov compression metaphor gives a widely applicable
picture of scientific knowledge, when it is restricted to {\it one of many} timescales
of natural phenomena: cf.~[15] and the LIGO story.

\smallskip

In the papers [23] and [21],  it was argued that brains actually
also use neural codes  allowing good compression
of relevant information.

\smallskip

 One set of arguments suggested that such a compression of,
 say, dictionary of the mother language in human brain can explain the well-known
 empirical observation,  Zipf's law.
 
 \smallskip
 This ``law'' (in fact, a keen and very general observation)  states that
 if one ranges lexemes in the order of their decreasing frequency of usage
 in a representative corpus of texts, then the product of lexeme frequency by
 the lexeme rank is approximately constant. In [23] it was argued
 that a good mathematical model of such behaviour is furnished
 by the L.~Levin's probability distribution, if one postulates
 that Zipf's ranking coincides with (an approximation to)  Kolmogorov's complexity
 ranking. 
 
 \smallskip
 The fact that Kolmogorov complexity in strict mathematical sense itself
 is not computable cannot refute this conjecture. In fact,
 successive approximations to Kolmogorov complexity ranking
 can be obtained by a version of the well-known ranking
 algorithm. 
 
 \smallskip
 Consider, for example, encoding and storing in the brain
 of the vocabulary of a mother language.
 We suggest that when a new lexeme is being encoded
 in a brain memory network, the length of this encoding
 (Zipf's ``effort'') is compared with lengths of previously encoded
 lexemes, and the lexeme acquires its temporary Zipf's rank.
 
  \smallskip
  
  Another set of arguments combined the discussion of neural
  encoding of stimulus spaces in [6] with suggestions
  of  [30] that dynamics in neural networks shows
  signatures of criticality.  ``Criticality'' here means that, within a certain
  statistical model of the relevant network, this dynamics happens
  near a phase transition regime. But it was discovered
  in [25] that search for good error-correcting (``noise-resistant'')
  codes generally involves activity near a phase transition curve,
  even though the relevant statistical model does not coincide
  with the one in  [30]: in fact, it again involves
  Kolmogorov complexity.

\smallskip

Stretching the metaphor further, we can also consider human
communication occurring in natural language in the same
light. A natural language message is usually treated as carrying
information. But it also can be treated as a program that runs in the
brain of the receiver and whose purpose is to create a certain
mind state in it. This interpretation is particularly interesting for
literary texts, especially poetry, because their purpose is not
conveying information, but rather imparting an emotional state to the
reader. It is customary to state that successful poetry {\it compresses} its
language and, consequently, if one wants to fully explicate the ``meaning''
of a good poem, an extensive prose text has to be written. So perhaps
the right way to conceptualize a great poem is to say that it represents  a maximally
Kolmogorov-compressed representation of the target mind state.

\smallskip

In the theoretical computer science, besides complexity as the length  of a 
shortest program, an important role is played by
various embodiments of the notion  ``length/time of computation.'' From this viewpoint,
 we are interested in minimizing time necessary for  producing
the output from an input. The most accomplished theory here led to the so called
``P/NP problem''. Roughly speaking, if {\it there exists} a computation of a function which requires time
polynomially  bounded by the length of input, can one also {\it find} this computation
using polynomially bounded time?

\smallskip
More precisely, in a model of the {\it universal} NP problem we consider
all Boolean polynomials  $F$  with arbitrary number of variables, and ask the question,
whether a given polynomial takes value 1 for some values of its arguments.

\smallskip

If the answer for $F$ is positive, this fact can be proved in polynomially 
bounded time (wrt the length of $F$) by starting
with an appropriate Boolean vector $x$ and then calculating the value  $F(x)=1$.
But can we {\it find this $x$} or else find {\it another proof} that $F$ takes value $1$
in polynomially bounded time? This is the P/NP problem the answer to which answer is not known.

\smallskip

What is relevant for our discussion here  is the fact that if we allow parallel
computations in our models, such as parallel computation of {\it all}  values
of any given Boolean polynomial by starting with all inputs of given length simultaneously, 
then the P/NP problem will obviously
have the answer P=NP. Thus, economy in computation time can be achieved
by allowing multiple parallelism.

\smallskip

This, in addition to program compression, might be another crucial mathematical idea that materializes in large
networks, both in brains and in civilizations.

\smallskip

Returning to the intuitive idea of compression, we want now to argue
that there is another type of compression which we will call here ``Darwinian compression.''

\medskip

{\bf Darwinian compression.} Charles Darwin's Beagle voyage was one of the defining
events in the development of human civilization because it has  radically changed 
our collective self-perception.

\smallskip

Narrowing our focus to see better his method from the viewpoint of its cognitive
characteristics, we can say that Darwin started with collecting  a vast database 
of living creatures. The contemporary ideology of data mining could suggest us
that his next step would be the  search for correlations in this database and 
discovery of various degrees of their possible interrelationships.
However, this kind of research was essentially done before Darwin: Carl Linn\'e
introduced the binary classification system (genus/species) and created
the principles of taxonomy that are still widely used.

\smallskip

Darwin's great breakthrough consisted in guessing how this diversity
could have occurred and what factors could determine the origin, development
and change of genera and species. The possibility to compress
his intuition in just two words, `` natural selection,'' motivated our
metaphor ``Darwinian compression''.

\smallskip

But in reality, one cannot rigorously derive, say,  the evolution theory from
genomics: all our attempts and arguments
are of vague qualitative nature, at best convincing us that the two sets of laws are compatible. 
A succinct and very expressive description of this baffling situation was given by
Svante P\"a\"abo in his book [26]: {\it `The dirty little secret of genomics is that we still know
next to nothing about how a genome translates into the particularities of a living and
breathing individual.' } Hence we cannot say which genomes would define
``the fittest'' individuals that, according to the Darwinian metaphor, have better chances for survival
and reproduction.

\smallskip

Attempts to fill this gap led to the development of  ``epigenetics,'' which is studying
factors and developmental processes that modify the activation of various genes without changing
the genetic code sequence of DNA: cf. [32].  Such epigenetic processes in a chromosome
can lead to the appearance of stably heritable phenotype traits, which then can play their
own specific roles in Darwinian evolution.

\smallskip

Another example of a scientific discovery of a similar cognitive type is the Periodic Table
of chemical elements (Mendeleev 1869) which embodied a compression
of a huge database of alchemical and later chemical observations, experiments, and guesses.

\smallskip

Both discoveries, evolution and periodic table, can be considered as a way of connecting various floors
of  scientific knowledge referring to various {\it space-time/complexity/ } scales.
 Each floor is governed by its ``laws'' in the sense
described above, which in principle should  be used to generate the laws of the next floor.

\smallskip

But, as in the case of Darwinian evolution, one cannot rigorously derive the periodic table
from the quantum theory of elementary particles and fields, and one cannot rigorously
derive, say, observable properties of water, ice, and steam from the position of  $H$ and $O$
in the periodic table.

\smallskip

 More precisely, quantitative 
theory of atoms of
the lightest elements consisting of a minimal number of elementary particles
might be accessible (with the help of modern computation resources),
but the whole structure of the table (including isotopes) and the very notion of molecules and their
``chemical'' properties, with its continuing extensions and ramifications all
the way up to DNA encoding, remain the ``upper-floor'' science, not really
reducible to the science ``one floor below.''

\smallskip

This is why we find so naive (and potentially dangerous) the claim by Chris Anderson,
Editor in Chief  of the ``Wired Magazine'', expressed in the title of the cover story 
``The End of Theory: The Data Deluge Makes the Scientific Method
Obsolete'' (summer 2008):

\smallskip

{\it The new availablility of huge amounts of data, along with statistical tools
to crunch these numbers, offers a whole new way of understanding the world.
Correlation supersedes causation, and science can advance even without coherent 
models, unified theories, or really any mechanical explanation at all. There's 
no reason to cling to our old ways.}

\smallskip

For more detailed arguments, cf. [22].

\medskip

{\bf Addendum.} Several weeks after completion and acceptance  of this article for 
publication, our attention was drawn to the beautiful book [33]. It presents the history
of humanity based upon the same metaphor as ours, but the book is much wider in scope
because it does not restrict itself to the study of only {\it cognitive} networks. The Great Silk Road
and Plato's Academy, Confucius and Martin Luther become routers and routes of the great web of civilizations.
\smallskip

We conclude this survey by the quotation from [33]:

\smallskip
`` [\dots ]   cultural evolution is Lamarckian, that is, acquired traits and skills can be passed on over
generations. Information -- how to speak a language or how to make people trust you -- is transmitted from brain to brain, from generation to generation, without the slow process of genetic mutation and natural selection. This accelerated pace
of cultural evolution made it possible for some groups of humankind to get the jump
on others and to destroy their structures and appropriate their resources. This does not
often happen in biological evolution
because it is slower: even the most complex creatures  evolve so slowly that others usually
have time to adapt. ''

\bigskip

\centerline{\bf References}

\medskip

[1] Yu.~ I.~Arshavsky. {\it Neurons versus Networks:
The interplay between individual neurons and neural networks in cognitive
functions.} The Neuroscientist, pp.~1--15, 2016.

\smallskip

[2]  M.~Borch--Jacobsen, S.~Shamsadani. {\it The Freud files. An inquiry
into the history of psychoanalysis.} Cambridge UP, 2012.

\smallskip

[3]  B.~J.~Copeland. {\it Turing's O--Machines, Penrose, Searle, and the Brain.}
Analysis, vol.~58, pp. 128--138, 1998.

\smallskip

[4]  B.~J.~Copeland. {\it Turing. Pioneer of the information age.}
Oxford UP, 2012

\smallskip
[5]  C.~Curto, V.~Itskov. {\it Cell groups reveal structure of stimulus space.} 
PLoS Computational Biology, vol.~4, issue 10, October 2008, 13 pp. (available online).

\smallskip

[6]  C.~Curto, V.~Itskov, A.~Veliz-Cuba, N.~Youngs. {\it The neural ring:
An algebraic tool for analysing the intrinsic structure of neural codes.}
Bull. Math.~Biology, 75(9), pp. 1571--1611, 2013.

\smallskip

[7]  {\it Space, time and number in the brain.} Ed.~by S.~Dehaine, E.~Brannon.
Elsevier Academic Press, 2011.
\smallskip

[8]  D.~Derdikman, E.~Moser. {\it A manifold of spatial maps in the brain.}
in: [7], pp.~41--57.

\smallskip

[9]  P.~F.~Ferrari, G.~Rizzolatti. {\it Mirror neuron research: the past and the future.}
Phil.~Transactions B, June, 2014.

\smallskip

[10]  D.~George, J.~Hawkins. {\it Towards a Mathematical Theory of Cortical
Micro--circuits.} PLoS Computational Biology, vol.~5, issue 10, October 2009,
26 pp.

\smallskip

[11]  J.~Hawkins with S.~Blakeslee. {\it On intelligence.}  Times books, NY 2004, 261 pp.

\smallskip

[12]  J.~Hawkins, D.~George and J.~Niemasik. {\it Sequence memory for prediction,
inference and behaviour.}  Phil.~Transactions of the Royal Society B, vol. 364, 2009, pp. 1203--1209.
\smallskip

[13]  J.~Hawkins, S.~Ahmad. {\it Why Neurons Have Thousands of Synapses,
a Theory of Sequence Memory in Neocortex.} Front. Neural Circuitsvol. vol 10, article 23, March 2016,
13 pp.

\smallskip

[14]  R.~Hersh. {\it Pluralism as Modelling and as Confusion.} (In this collection).

\smallskip

[15] G.~${}^{\prime}$t Hooft, St.~Vandoren. {\it Time in powers of ten. Natural
phenomena and their timescales.} World Scientific, 2014.

\smallskip

[16]  P.~Indefrey, W.~J.~M.~Levelt. {\it The spatial and temporal signatures of word production
components.} Cognition, 92, 2004, pp.~101--144

\smallskip

[17] Br.~Knight, D.~Manin, L.~Sirovich. {\it Dynamical models of interacting neuron populations
in visual cortex.} Robot Cybern., v.~54, 1996, pp. 4--8.

\smallskip

[18] Yu.~I. Manin. {\it Mathematics as Metaphor. Selected essays.} American Math.~Society, 2007.

\smallskip

[19]  Yu.~I. Manin. {\it  Neural codes and homotopy types: mathematical models of
place field recognition.}   Moscow Math. Journal, vol. 15, Oct.--Dec. 2015,  pp. 741--748 . Preprint
arXiv:1501.00897 

\smallskip

[20]  Yu.~I.~Manin. {\it Cognition and Complexity.} In: M. Burgin, C.S. Calude (eds.). Information and Complexity.
World Scientific Series in Information Studies, 2016, pp. 344--357.

\smallskip

[21] Yu.~I.~Manin. {\it  Error--correcting codes and neural networks.}  Selecta Math. New. Ser. (2016).
 doi:10.1007/s00029-016-0284-4 . 10 pp.

\smallskip

[22]  Yu.~I.~Manin. {\it Kolmogorov complexity as a hidden factor of scientific discourse: from
Newton's law to data mining.} In:  "Complexity and Analogy in Science: Theoretical, Methodological
and Epistemological Aspects", Proceedings of the Plenary session of Pontifical Ac.~Sci.,  November 5--7,
2012.} Libreria Editrice Vaticana, 2015. Preprint arXiv:1301.0081.

\smallskip

[23]  Yu.~I.~Manin. {\it Zipf's law and L.~Levin's probability distributions.} Functional Analysis and its Applications,
vol. 48, no. 2, 2014. DOI 10.107/s10688-014-0052-1.
Preprint arXiv:
1301.0427

\smallskip

[24]  Yu.~I.~Manin. {\it De Novo Artistic Activity, Origins of Logograms, and Mathematical
Intuition.} In: Art in the Life of Mathematicians, Ed. Anna Kepes Szemer\'edi, AMS, 2015 pp. 187--208.
\smallskip

[25]   Yu.~I.~Manin, M.~Marcolli. {\it Error--correcting codes and phase transitions.}
 Mathematics in Computer Science, vol.~5 (2011), 133--170.
Preprint mat.QA/0910.5135

\smallskip 

[26] Svante P\"a\"abo. {\it  Neanderthal Man: In Search of Lost Genomes.} 
Basic Books, NY, 2014, 275 pp.

\smallskip

[27]  R.~Q.~Quiroga. {\it Concept cells: the building blocks of declarative memory
functions.} Nature reviews $|$ Neuroscience, vol. 13, Aug. 2012, pp.~587--597.

\smallskip

[28]   R.~Q.~Quiroga, L.~Reddy, G.~Kreiman, C.~Koch, I.~Fried.
{\it Invariant visual representation by single neurons in in
the human brain.} Nature 435, 2005, pp. 1102--1107.

\smallskip

[29]  D.~A.~Sakharov.  {\it Cognitive pattern generators.} Lecture at the 6th International
conference on cognitive science, 2014.

\smallskip

[30]  G.~Tka\v{c}ik, T.~Mora, O.~Marre, D.~Amodei, S.~Palmer, M.~Berry, W.~Bialek.
 {\it Thermodynamics and signatures of criticality in a network of neurons.}
Proc.~Nat.~Ac.~Sci, 112(37):11, 2015, pp. 508--517.

\smallskip

[31] H.~Zimmerman. {\it OSI Reference Model -- the ISO model of architecture
for open systems interconnection.} IEEE Transactions on Communications,
28 (4), 1980, pp. 425--432

\smallskip

[OSI1] https://en.wikipedia.org/wiki/OSI${}_{-}$mode

[OSI2] https://en.wikipedia.org/wiki/Wayback${}_{-}$Machine

[OSI3] https://en.wikipedia.org/wiki/Knowledge${}_{-}$Graph

\smallskip
[32] S.~L.~Berger, T.~Couzarides, R.~Shiekhattar, A.~Shilatifard.
{\it An operational definition of epigenetics.} Genes Dev., 23(7), 2009, pp. 781--783.

\smallskip

[33]  J.~R.~McNeill and William H.~McNeill.
{\it The Human Web: a Bird's-Eye View of World History.}
Norton, 2003. xviii + 350 pp.

\enddocument